\documentclass[final,onefignum,onetabnum]{siamart190516}

\usepackage{braket,amsfonts}
\usepackage{boldline, makecell, booktabs}
\usepackage{mathabx}
\usepackage{multirow}
\usepackage{array}

\usepackage[caption=false]{subfig}
\usepackage{pgfplots}

\usepackage{cleveref}
\usepackage{hyperref}

\newsiamthm{claim}{Claim}
\newsiamremark{remark}{Remark}
\newsiamremark{hypothesis}{Hypothesis}
\crefname{hypothesis}{Hypothesis}{Hypotheses}

\usepackage{algorithmic}

\usepackage{graphicx,epstopdf}

\Crefname{ALC@unique}{Line}{Lines}

\usepackage{amsopn}

\makeatletter
\DeclareRobustCommand{\cev}[1]{%
  \mathpalette\do@cev{#1}%
}
\newcommand{\do@cev}[2]{%
  \fix@cev{#1}{+}%
  \reflectbox{$\m@th#1\vec{\reflectbox{$\fix@cev{#1}{-}\m@th#1#2\fix@cev{#1}{+}$}}$}%
  \fix@cev{#1}{-}%
}
\newcommand{\fix@cev}[2]{%
  \ifx#1\displaystyle
    \mkern#23mu
  \else
    \ifx#1\textstyle
      \mkern#23mu
    \else
      \ifx#1\scriptstyle
        \mkern#22mu
      \else
        \mkern#22mu
      \fi
    \fi
  \fi
}

\usepackage{xspace}
\usepackage{bold-extra}
\usepackage[most]{tcolorbox}

\colorlet{texcscolor}{blue!50!black}
\colorlet{texemcolor}{red!70!black}
\colorlet{texpreamble}{red!70!black}
\colorlet{codebackground}{black!25!white!25}

\lstdefinestyle{siamlatex}{%
  style=tcblatex,
  texcsstyle=*\color{texcscolor},
  texcsstyle=[2]\color{texemcolor},
  keywordstyle=[2]\color{texemcolor},
  moretexcs={cref,Cref,maketitle,mathcal,text,headers,email,url},
}

\tcbset{%
  colframe=black!75!white!75,
  coltitle=white,
  colback=codebackground, 
  colbacklower=white, 
  fonttitle=\bfseries,
  arc=0pt,outer arc=0pt,
  top=1pt,bottom=1pt,left=1mm,right=1mm,middle=1mm,boxsep=1mm,
  leftrule=0.3mm,rightrule=0.3mm,toprule=0.3mm,bottomrule=0.3mm,
  listing options={style=siamlatex}
}

\newtcblisting[use counter=example]{example}[2][]{%
  title={Example~\thetcbcounter: #2},#1}

\newtcbinputlisting[use counter=example]{\examplefile}[3][]{%
  title={Example~\thetcbcounter: #2},listing file={#3},#1}

\DeclareTotalTCBox{\code}{ v O{} }
{ 
  fontupper=\ttfamily\color{black},
  nobeforeafter,
  tcbox raise base,
  colback=codebackground,colframe=white,
  top=0pt,bottom=0pt,left=0mm,right=0mm,
  leftrule=0pt,rightrule=0pt,toprule=0mm,bottomrule=0mm,
  boxsep=0.5mm,
  #2}{#1}

\patchcmd\newpage{\vfil}{}{}{}
\begin{tcbverbatimwrite}{tmp_\jobname_header.tex}
\title{Diffusion-Model-Assisted Supervised Learning of Generative Models for Density Estimation\thanks{{\bf Notice}:  This manuscript has been authored by UT-Battelle, LLC, under contract DE-AC05-00OR22725 with the US Department of Energy (DOE). The US government retains and the publisher, by accepting the article for publication, acknowledges that the US government retains a nonexclusive, paid-up, irrevocable, worldwide license to publish or reproduce the published form of this manuscript, or allow others to do so, for US government purposes. DOE will provide public access to these results of federally sponsored research in accordance with the DOE Public Access Plan.
}}

\author{Yanfang Liu\thanks{Computer Science and Mathematics Division, Oak Ridge National Laboratory, Oak Ridge, TN 37831, USA.}
\and Minglei Yang\thanks{Fusion Energy Science Division, Oak Ridge National Laboratory, Oak Ridge, TN 37831, USA.}
\and Zezhong Zhang\footnotemark[2]
\and Feng Bao\thanks{Department of Mathematics, Florida State University, Tallahassee, FL 32306, USA.}
\and Yanzhao Cao\thanks{Department of Mathematics and Statistics, Auburn University, Auburn, AL 36849, USA.}
\and Guannan Zhang\footnotemark[2]
}

\headers{Supervised Learning of Probabilistic GMs for Density Estimation}{Y.~Liu, M.~Yang, Z.~Zhang, F.~Bao, Y.~Cao, G.~Zhang}
\end{tcbverbatimwrite}
\title{Diffusion-Model-Assisted Supervised Learning of Generative Models for Density Estimation\thanks{{\bf Notice}:  This manuscript has been authored by UT-Battelle, LLC, under contract DE-AC05-00OR22725 with the US Department of Energy (DOE). The US government retains and the publisher, by accepting the article for publication, acknowledges that the US government retains a nonexclusive, paid-up, irrevocable, worldwide license to publish or reproduce the published form of this manuscript, or allow others to do so, for US government purposes. DOE will provide public access to these results of federally sponsored research in accordance with the DOE Public Access Plan.
}}

\author{Yanfang Liu\thanks{Computer Science and Mathematics Division, Oak Ridge National Laboratory, Oak Ridge, TN 37831, USA.}
\and Minglei Yang\thanks{Fusion Energy Science Division, Oak Ridge National Laboratory, Oak Ridge, TN 37831, USA.}
\and Zezhong Zhang\footnotemark[2]
\and Feng Bao\thanks{Department of Mathematics, Florida State University, Tallahassee, FL 32306, USA.}
\and Yanzhao Cao\thanks{Department of Mathematics and Statistics, Auburn University, Auburn, AL 36849, USA.}
\and Guannan Zhang\footnotemark[2]
}

\headers{Supervised Learning of Probabilistic GMs for Density Estimation}{Y.~Liu, M.~Yang, Z.~Zhang, F.~Bao, Y.~Cao, G.~Zhang}

\begin{document}
\maketitle

\begin{tcbverbatimwrite}{tmp_\jobname_abstract.tex}
\begin{abstract}
We present a supervised learning framework of training generative models for density estimation. Generative models, including generative adversarial networks, normalizing flows, variational auto-encoders, are usually considered as unsupervised learning models, because labeled data are usually unavailable for training. Despite the success of the generative models, there are several issues with the unsupervised training, e.g., requirement of reversible architectures, vanishing gradients, and training instability. 
To enable supervised learning in generative models, we utilize the score-based diffusion model to generate labeled data. Unlike existing diffusion models that train neural networks to learn the score function, we develop a training-free score estimation method. This approach uses mini-batch-based Monte Carlo estimators to directly approximate the score function at any spatial-temporal location in solving an ordinary differential equation (ODE), corresponding to the reverse-time stochastic differential equation (SDE). This approach can offer both high accuracy and substantial time savings in neural network training. Once the labeled data are generated, we can train a simple fully connected neural network to learn the generative model in the supervised manner. 
Compared with existing normalizing flow models, our method does not require to use reversible neural networks and avoids the computation of the Jacobian matrix. Compared with existing diffusion models, our method does not need to solve the reverse-time SDE to generate new samples. As a result,  the sampling efficiency is significantly improved. We demonstrate the performance of our method by applying it to  a set of 2D datasets as well as real data from the UCI repository.
\end{abstract}


\end{tcbverbatimwrite}
\begin{abstract}
We present a supervised learning framework of training generative models for density estimation. Generative models, including generative adversarial networks, normalizing flows, variational auto-encoders, are usually considered as unsupervised learning models, because labeled data are usually unavailable for training. Despite the success of the generative models, there are several issues with the unsupervised training, e.g., requirement of reversible architectures, vanishing gradients, and training instability.
To enable supervised learning in generative models, we utilize the score-based diffusion model to generate labeled data. Unlike existing diffusion models that train neural networks to learn the score function, we develop a training-free score estimation method. This approach uses mini-batch-based Monte Carlo estimators to directly approximate the score function at any spatial-temporal location in solving an ordinary differential equation (ODE), corresponding to the reverse-time stochastic differential equation (SDE). This approach can offer both high accuracy and substantial time savings in neural network training. Once the labeled data are generated, we can train a simple fully connected neural network to learn the generative model in the supervised manner.
Compared with existing normalizing flow models, our method does not require to use reversible neural networks and avoids the computation of the Jacobian matrix. Compared with existing diffusion models, our method does not need to solve the reverse-time SDE to generate new samples. As a result,  the sampling efficiency is significantly improved. We demonstrate the performance of our method by applying it to  a set of 2D datasets as well as real data from the UCI repository.
\end{abstract}




\section{Introduction}\label{sec:intro}
Density estimation involves the approximation of the probability density function of a given set of observation data points. The overarching goal is to characterize the underlying structure of the observation data. Generative models belong to a class of machine learning (ML) models designed to model the underlying probability distribution of a given dataset, enabling the generation of new samples that are statistically similar to the original data. Many methods for generative models 
have been proposed over the past decade, including variational auto-encoders (VAEs) \cite{DBLP:journals/corr/KingmaW13}, generative adversarial networks (GANs) \cite{NIPS2014_5ca3e9b1}, normalizing flows \cite{kobyzev2020normalizing}, and diffusion models \cite{10.1145/3626235}.
Generative models have been successfully used in a variety of applications, including image synthesis \cite{NEURIPS2019_3001ef25,DBLP:conf/eccv/CaiYAHBSH20,DBLP:journals/jmlr/HoSCFNS22,DBLP:conf/iclr/MengHSSWZE22,NIPS2014_5ca3e9b1}, image denoising \cite{DBLP:conf/iccv/LuoH21,NEURIPS2020_4c5bcfec,DBLP:conf/icml/Sohl-DicksteinW15,DBLP:conf/cvpr/LedigTHCCAATTWS17}, anomaly detection \cite{DBLP:conf/ipmi/SchleglSWSL17,papamakarios2017masked} and natural language processing \cite{DBLP:conf/nips/AustinJHTB21,DBLP:conf/nips/HoogeboomNJFW21,DBLP:journals/corr/abs-2205-14217,DBLP:conf/iclr/SavinovCBEO22,DBLP:conf/icml/YuXMJPGZZW22,DBLP:conf/emnlp/MaZLNH19}. The key idea behind recent generative models is to exploit the superior expressive power and flexibility of deep neural networks to detect and characterize complicated geometries embedded in the possibly high-dimensional observational data. 

Generative models for density estimation are usually considered as unsupervised learning, primarily due to  the absence of labeled data. Various unsupervised loss functions have been defined to train the underlying neural network models in generative models, including adversarial loss for GANs \cite{NIPS2014_5ca3e9b1}, the maximum likelihood loss for normalizing flows \cite{kobyzev2020normalizing}, and the score matching losses for diffusion models 
\cite{JMLR:v6:hyvarinen05a,10.1162/NECO_a_00142,pmlr-v115-song20a}. Despite the success of the generative models, there are several issues resulted from the unsupervised training approach. For example, the training of GANs may suffer from mode collapse, vanishing gradients, and training instability \cite{DBLP:conf/nips/SalimansGZCRCC16}. The maximum likelihood loss used in normalizing flows requires efficient computation of the determinant of the Jacobian matrix, which places severe restrictions on networks' architectures. If labeled data can be created based on the observational data without complicated training, the generator in the generative models, e.g., the decoder in VAEs or normalizing flows, can be trained in a supervised manner, which can circumvent  the issues in unsupervised training. 

In this work, we propose a supervised learning framework of training generative models for density estimation. The key idea is to use the score-based diffusion model to generate labeled data and then use the simple mean squared loss (MSE) to train the generative model. 
A diffusion model can transport the standard Gaussian distribution to a complex target data distribution through a reverse-time diffusion process in the form of a stochastic differential equation (SDE), and the score function in the drift term guides the reverse-time SDE towards the data distribution. Since the standard Gaussian distribution is independent of the target data distribution, the information of the data distribution is fully stored in the score function. Thus, we use the reverse-time SDE to generate the labeled data.
Unlike existing diffusion models that train neural networks to learn the score function \cite{song2021scorebased,bao2023scorebased,pmlr-v180-shi22a}, 
we develop a training-free score estimation that uses mini-batch-based Monte Carlo estimators to directly approximate the score function at any spatial-temporal location in solving the reverse-time SDE. 
Numerical examples in Section \ref{sec:ex} demonstrate that the training-free score estimation approach offers sufficient accuracy and save significant computing cost on training neural networks in the meantime. Once the labeled data are generated, we can train a simple fully connected neural network to learn the generator in the supervised manner. 
Compared with existing normalizing flow models, our method does not require to use reversible neural networks and avoids the computation of the Jacobian matrix. Compared with existing diffusion models, our method does not need to solve the reverse-time SDE to generate new samples. This way, the sampling efficiency is significantly improved.

The rest of this paper is organized as follows. In Section \ref{sec:prob}, we briefly introduce the density estimation  problem under consideration. In Section \ref{sec:method}, we provide a comprehensive discussion of the proposed method. Finally in  Section \ref{sec:ex}, we demonstrate the performance of our method by applying it to a set of 2D datasets as well as real data from the UCI repository.

\section{Problem setting}\label{sec:prob}
We are interested in learning how to generate unrestricted number of  samples of a target $d$-dimensional random variable, denoted by
\begin{equation}\label{eq:target}
    X \in \mathbb{R}^d \text{ and } X \sim p(x),
\end{equation}
where $p$ is the probability density function of $X$. 
We aim to achieve this from a given dataset, denoted by $\mathcal{X} = \{x_1, x_2, \ldots, x_J\} \subset \mathbb{R}^d$, and from  its probability density function (PDF) $p$. To this end, we aim at building a parameterized generative model, denoted by
\begin{equation}\label{eq:transport}
    X = F(Y; \theta)\; \text{ with }\; Y \in \mathbb{R}^d,
\end{equation}
which is a transport model that maps a reference variable $Y$ (usually following a standard Gaussian distribution) to the target random variable $X$. Once the optimal value for the parameter $\theta$ is obtained by training the model, we can generate samples of $X$ by drawing samples of $Y$ and pushing them through the trained generative model. 

This problem has been extensively studied in the machine learning community using normalizing flow models \cite{kobyzev2020normalizing,rezende2015variational,creswell2018generative,papamakarios2017masked, dinh2016density, grathwohl2018ffjord,GUO2022111202}, where $F(y; \theta)$ is defined as a bijective function. One major challenge for training the generative model $F(y; \theta)$ is the lack of labeled training data, i.e., there is no corresponding sample of $Y$ for each sample $x_j \in \mathcal{X}$.  As a result,  the model $F(y; \theta)$ cannot be trained via the simplest supervised learning using the mean square error (MSE) loss. Instead, an indirect loss defined by the change of variables formula, i.e., $p(x) = p(F^{-1}(x))|\det \mathbf{D}(F^{-1}(x))|$, is extensively used to train normalizing flow models in an unsupervised manner. The drawback of this training approach is that specially designed reversible architectures for $F(y; \theta)$ need to be used to efficiently perform backpropagation through the computation of $|\det \mathbf{D}(F^{-1}(x))|$. In this paper, we address this challenge by using score-based diffusion models to generate labeled data such that the generative model $F(y;\theta)$ can be trained in a supervised manner and $F^{-1}$ is no longer needed in the training. 

\section{Methodology}\label{sec:method}
In this section, we present in details the proposed method.  We briefly review the score-based diffusion model in Section \ref{sec:diff}. In Section \ref{sec:exact_score}, we introduce the training-free score estimation approach that is needed for generating the labeled data. The main algorithm for the supervised learning of the generative model $F(y;\theta)$ is presented in Section \ref{sec:learning}.

\subsection{The score-based diffusion model}\label{sec:diff}
In this subsection, we will provide a brief overview of score-based diffusion models (see \cite{song2021scorebased,pmlr-v115-song20a} for more details). The model under consideration consists of a forward SDE and a reverse-time SDE defined in a standard temporal domain $[0,1]$. The forward SDE, defined by 
\begin{equation}\label{eq:forward}
d Z_t = b(t) Z_{t} dt + \sigma(t) dW_t\; \text{ with }\; Z_0 = X \text{ and } Z_1 = Y,
\end{equation}
is used to map the target random variable $X$ (i.e., the initial state $Z_0$) to the standard Gaussian random variable $Y \sim \mathcal{N}(0, \mathbf{I}_d)$ (i.e., the terminal state $Z_1$). There is a number of choices for the drift and diffusion coefficients in Eq.~\eqref{eq:forward} to ensure that the terminal state $Z_1$ follows $\mathcal{N}(0, \mathbf{I}_d)$ (see  \cite{song2021scorebased,10.1162/NECO_a_00142,NEURIPS2020_4c5bcfec,lu2022dpmsolver} for details). In this work, we choose $b(t)$ and $\sigma(t)$ in Eq.~\eqref{eq:forward} as 
\begin{equation}\label{eq:cof}
\begin{aligned}
b(t) = \frac{{\rm d} \log \alpha_t}{{\rm d} t} \;\;\; \text{ and }\;\;\; \sigma^2(t) = \frac{{\rm d} \beta_t^2}{{\rm d}t} - 2 \frac{{\rm d}\log \alpha_t}{{\rm d}t} \beta_t^2,
\end{aligned}
\end{equation}
where the two processes $\alpha_t$ and $\beta_t$ are defined by
\begin{equation}\label{eq:ab}
\alpha_t = 1-t, \;\; \beta_t^2 = t \;\; \text{ for } \;\; t \in [0,1].
\end{equation}
Because the forward SDE in \eqref{eq:forward}
is linear and driven by an additive noise, the conditional probability density function $Q(Z_t | Z_0)$ for any fixed $Z_0$ is a Gaussian. In fact, 
\begin{equation}\label{eq:gauss}
Q(Z_t | Z_0) = \mathcal{N}(\alpha_t Z_0, \beta_t^2 \mathbf{I}_d), 
\end{equation}
which leads to $Q(Z_1 | Z_0) = \mathcal{N}(0, \mathbf{I}_d)$. The corresponding reverse-time SDE is defined by
\begin{equation}\label{DM:RSDE}
d{Z}_t = \left[ b(t){Z}_t - \sigma^2(t) S(Z_t, t)\right] dt + \sigma(t) d\cev{W}_t\; \text{ with }\; Z_0 = X \text{ and } Z_1 = Y,
\end{equation}
where $\cev{W}_t$ is the backward Brownian motion and $S(Z_t, t)$ is the {score function}. If the score function is defined by 
\begin{equation}\label{eq:exact_score}
 S(Z_{t}, t) := \nabla_z \log Q({Z}_t),
\end{equation}
where $Q(Z_t)$ is the probability density function of $Z_t$ in Eq.~\eqref{eq:forward}, then the reverse-time SDE maps the standard Gaussian random variable $Y$ to the target random variable $X$.  Therefore, if we can evaluate the score function for any $(Z_t,t)$, then we can directly generate samples of $X$ by pushing samples of $Y$ through the reverse-time SDE. Thus, the central task in training diffusion models is appropriately approximating  the score function. There are several established approaches,  including score matching \cite{JMLR:v6:hyvarinen05a}, 
denoising score matching \cite{10.1162/NECO_a_00142}, and sliced score matching \cite{pmlr-v115-song20a}, etc. 
Recent advances in diffusion models focus on explorations of  using neural networks to approximate the score function. 
Compared to the normalizing flow models, a notable  drawback of learning the score function is its  inefficiency in sampling since  generating one sample of $X$ requires solving the reverse-time SDE. To alleviate this challenge, we use the score-based diffusion model as a data labeling approach, which will enable supervised learning of the generative model of interest.

\subsection{Training-free score estimation}\label{sec:exact_score}
We now derive the analytical formula of the score function and its approximation scheme. Substituting $Q(Z_t)= \int_{\mathbb{R}^d} Q(Z_t, Z_0) dZ_0= \int_{\mathbb{R}^d} Q({Z}_t | Z_0) Q(
Z_0)dZ_0$ into Eq.~\eqref{eq:exact_score} and exploiting the fact in Eq.~\eqref{eq:gauss}, we can rewrite the score function as 
\begin{equation}\label{eq:score11}
\begin{aligned}
S(Z_{t}, t) & = \nabla_z \log \left(\int_{\mathbb{R}^d} Q({Z}_t | Z_0) Q(
Z_0) dZ_0\right)\\
& = \frac{1}{\int_{\mathbb{R}^d} Q({Z}_t | Z'_0) Q(
Z'_0) dZ'_0}   \int_{\mathbb{R}^d}  - \frac{Z_t - \alpha_t Z_0}{\beta^2_t}Q({Z}_t | Z_0) Q(
Z_0) dZ_0\\
& =  \int_{\mathbb{R}^d}  - \frac{Z_t- \alpha_t Z_0}{\beta^2_t} w_t({Z}_t,  Z_0)  Q(Z_0)dZ_0,\\
\end{aligned}
\end{equation}
where the weight function $w_t({Z}_t,  Z_0)$ is defined by
\begin{equation}\label{eq:weight}
w_t({Z}_t,  Z_0) :=  \frac{ Q({Z}_t | Z_0) }{\int_{\mathbb{R}^d} Q({Z}_t | Z'_0) Q(
Z'_0) dZ'_0},
\end{equation}
satisfying that $\int_{\mathbb{R}^d}w_t({Z}_t,  Z_0) Q(Z_0)dZ_0 = 1$. 

The integrals/expectations in Eq.~\eqref{eq:score11} can be approximated by Monte Carlo estimators using the available samples in $\mathcal{X} = \{x_1, x_2, \ldots, x_J\} \subset \mathbb{R}^d$. According to the definition of the reverse-time SDE in Eq.~\eqref{DM:RSDE}, the samples in $\mathcal{X}$ are also samples from $Q(Z_0)$. Thus, the integral in Eq.~\eqref{eq:score11} can be estimated by 
\begin{equation}\label{eq:MC}
S(Z_t, t) \approx \bar{S}(Z_t, t) :=  \sum_{n=1}^{N} - \frac{Z_t - \alpha_t x_{j_n}}{\beta^2_t} \bar{w}_t({Z_t},  x_{j_n}), 
\end{equation}
using a mini-batch of the dataset $\mathcal{X}$ with batch size $N \le J$, denoted by $\{x_{j_n}\}_{n=1}^N$, where the weight $w_t({Z_t},x_{j_n})$ is calculated by
\begin{equation}\label{eq:weight_app}
w_t({Z_t}, x_{j_n}) \approx  \bar{w}_t({Z_t}, x_{j_n}) := \frac{Q(Z_t |  x_{j_n}) }{\sum_{m=1}^{N} Q(Z_t|  x_{j_{m}})},
\end{equation}
and $Q(Z_t |  x_{j_n})$ is the Gaussian distribution given in Eq.~\eqref{eq:gauss}. 
This means $w_t({Z_t},  Z_0)$ can be estimated by the normalized probability density values 
$\{Q(Z_t|x_{j_n})\}_{n=1}^N$. In practice, the size of the mini-batch $\{x_{j_n}\}_{n=1}^N$ can be flexibly adjusted to balance the efficiency and accuracy.

\subsection{Supervised learning of the generative model}\label{sec:learning}
\begin{figure}[h!]
    \centering
    \includegraphics[width=0.8\textwidth]{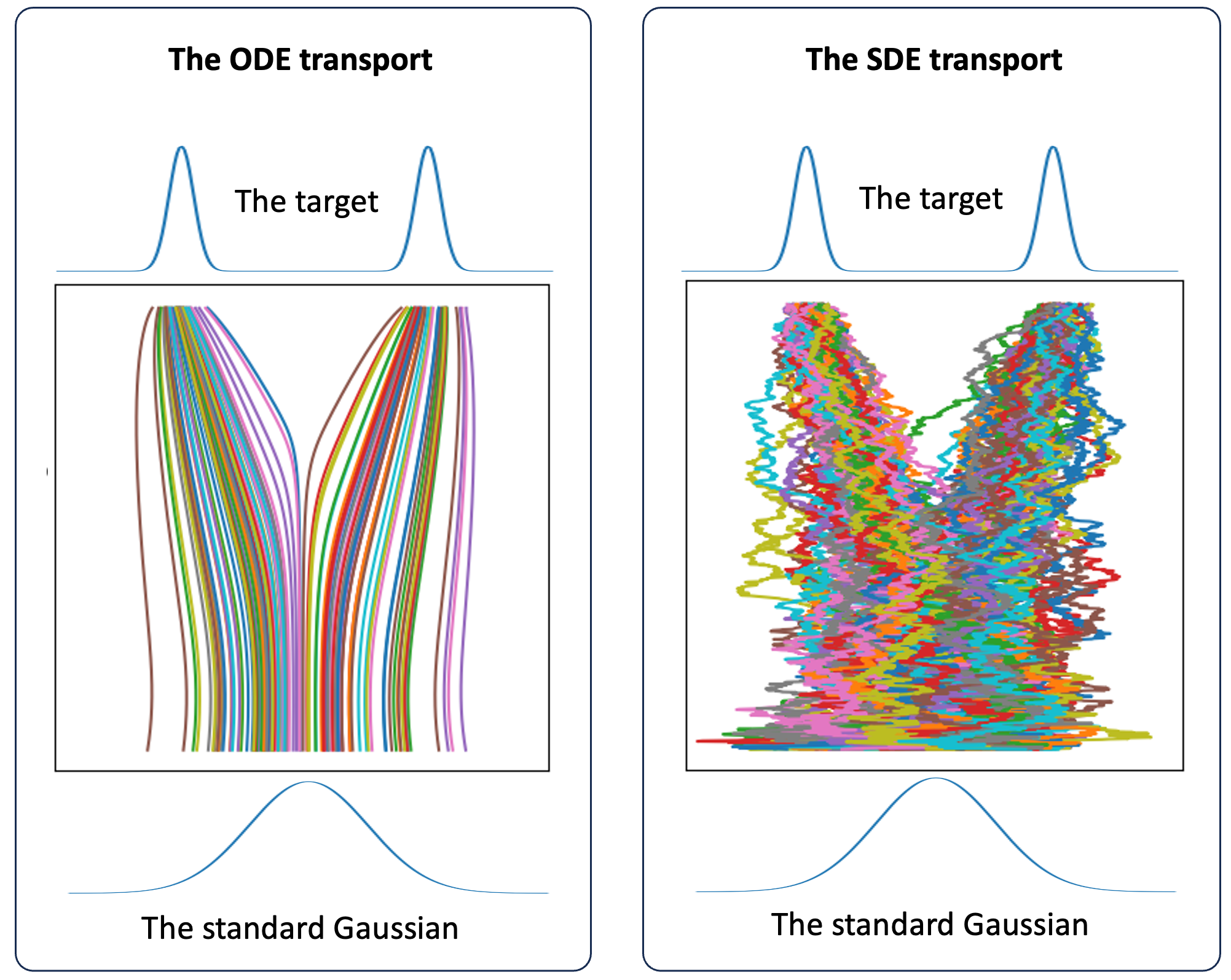}
    \caption{Illustration of the trajectories of the ODE model in Eq.~\eqref{eq:ode} and the SDE in Eq.~\eqref{DM:RSDE} using a simple one-dimensional example. We observe that the ODE model creates a much smoother function relationship between the initial state and the terminal state, which indicates that the ODE model is more suitable for generating the labeled data for supervised learning of the generative model of interest.}
    \label{fig:illustration}
    \vspace{-0.cm}
\end{figure}

In this subsection, we describe how to use the score approximation scheme given in Section \ref{sec:exact_score} to generate labeled data and use such data to train the generative model of interest. Due to the stochastic nature of the reverse-time SDE in Eq.~\eqref{DM:RSDE}, the relationship between the initial state $Z_0$ and the terminal state $Z_1$ is not deterministic or smooth, as shown in Figure \ref{fig:illustration}. Thus, we cannot directly use Eq.~\eqref{DM:RSDE} to generate labeled data. Instead, we use the corresponding ordinary differential equation (ODE), defined by
\begin{equation}\label{eq:ode}
    d{Z}_t = \left[ b(t){Z}_t - \frac{1}{2}\sigma^2(t) S(Z_t, t)\right] dt\; \text{ with }\; Z_0 = X \text{ and } Z_1 = Y,
\end{equation}
whose trajectories share the same marginal probability density functions as the reverse-time SDE in Eq.~\eqref{DM:RSDE}. An illustration of the trajectories of the SDE and the ODE is given in Figure \ref{fig:illustration}. We observe that ODE defines a much smoother function relationship between the initial state and the terminal state than that defined by the SDE. Thus, we use the ODE in Eq.~\eqref{eq:ode} to generate labeled data. 

Specifically, we first draw $M$ random samples of $Y$, denoted by $\mathcal{Y} = \{y_1, \ldots, y_M\}$ from the standard Gaussian distribution. For $m=1, \ldots,M$, we solve the ODE in Eq.~\eqref{eq:ode} from $t = 1$ to $t=0$ and collect the state $Z_0 | y_m$, where the score function is computed using Eq.~\eqref{eq:MC}, Eq.~\eqref{eq:weight_app}, and the dataset $\mathcal{X} = \{x_1, \ldots, x_J\}$. The labeled training dataset is denoted by 
\begin{equation}\label{eq:label}
    \mathcal{D}_{\rm train} := \{ (y_m, x_m): x_m = Z_0|y_m, \; \text{ for }\; m = 1, \ldots, M\},
\end{equation}
where $x_m$ is obtained by solving the ODE in Eq.~\eqref{eq:ode}. We note that the $x_m$'s in $\mathcal{D}_{\rm train}$ 
may not belong to $\mathcal{X}$ and $M$ could be arbitrarily large. After obtaining $\mathcal{D}_{\rm train}$ we can use it to train the generative model $F(y; \theta)$ in Eq.~\eqref{eq:transport} using supervised learning with the MSE loss.

\vspace{-.0cm}
\noindent\makebox[\linewidth]{\rule{\textwidth}{0.5pt}}\\
\vspace{-0.5cm}
\newline {\bf Algorithm 1: supervised learning of generative models}\vspace{-0.2cm} \\ 
\noindent\makebox[\linewidth]{\rule{\textwidth}{0.5pt}}
\vspace{-0.3cm}
\newline1:\, {\bf Input}: observation data $\mathcal{X}$, diffusion coefficient $\sigma(t)$, and drift coefficient $b(t)$;
\newline2:\, {\bf Output}: the trained generative model $F(y;\theta)$;
\vspace{0.15cm}
\newline3: Draw $M$ samples $\mathcal{Y} = \{y_1, \ldots, y_M\}$ from the standard Gaussian distribution;
\vspace{0.05cm}
\newline4: \,{\bf for} $m = 1, \ldots, M$
\vspace{0.1cm}
\newline5: \qquad Solve the ODE in Eq.~\eqref{eq:ode} with the score function estimated by
Eq.~\eqref{eq:MC} and Eq.~\eqref{eq:weight_app};
\vspace{0.1cm}
\newline6: \qquad Define a pair of labeled data $(y_m, x_m)$ where $y_m = Z_1$ and $x_m = Z_0$ in Eq.~\eqref{eq:ode};
\vspace{0.1cm}
\newline7: {\bf end}\vspace{-0.1cm} 
\vspace{0.15cm}
\newline8: Train the generative model $x = F(y;\theta)$ with the MSE loss.\vspace{-0.1cm} \\
\noindent\makebox[\linewidth]{\rule{\textwidth}{0.5pt}}\\
\vspace{-0.2cm}

Our method is summarized in Algorithm 1. Compared to the existing normalizing flow models and diffusion models, our method has two significant advantages  in performing density estimation tasks. First, it does not require to know $F^{-1}(x; \theta)$, hence it does not require  the computation of $|\det \mathbf{D}(F^{-1}(x))|$ in the training process.  This enables us to use simpler neural 
network architectures to define $F(y;\theta)$, resulting in  a more straightforward  training procedure compared to the training of a normalizing flow model.
Second, after $F(y; \theta)$ is trained, our method does not require solving  the reverse-time SDE or ODE to generate samples of $X$.  As a result, it significantly enhances the sampling efficiency in comparison with   the diffusion model.

\section{Numerical Experiments}\label{sec:ex}
We demonstrate the performance of the proposed method on several benchmark problems for density estimation. 
 To solve the reverse-time ODE for generating the labeled data,  we use theexplicit Euler scheme. 
The generative model $F(y;\theta)$ is defined by a fully-connected feed-forward neural network. 
Our method is implemented in Pytorch with GPU acceleration enabled. The source code is publicly available at \url{https://github.com/mlmathphy/supervised_generative_model}. The numerical results in this section can be precisely reproduced using the code on Github.

\subsection{Density Estimation on Toy 2D Data}\label{sec:2d}
\begin{figure}[h!]
    \centering
    \includegraphics[width=0.85\textwidth]{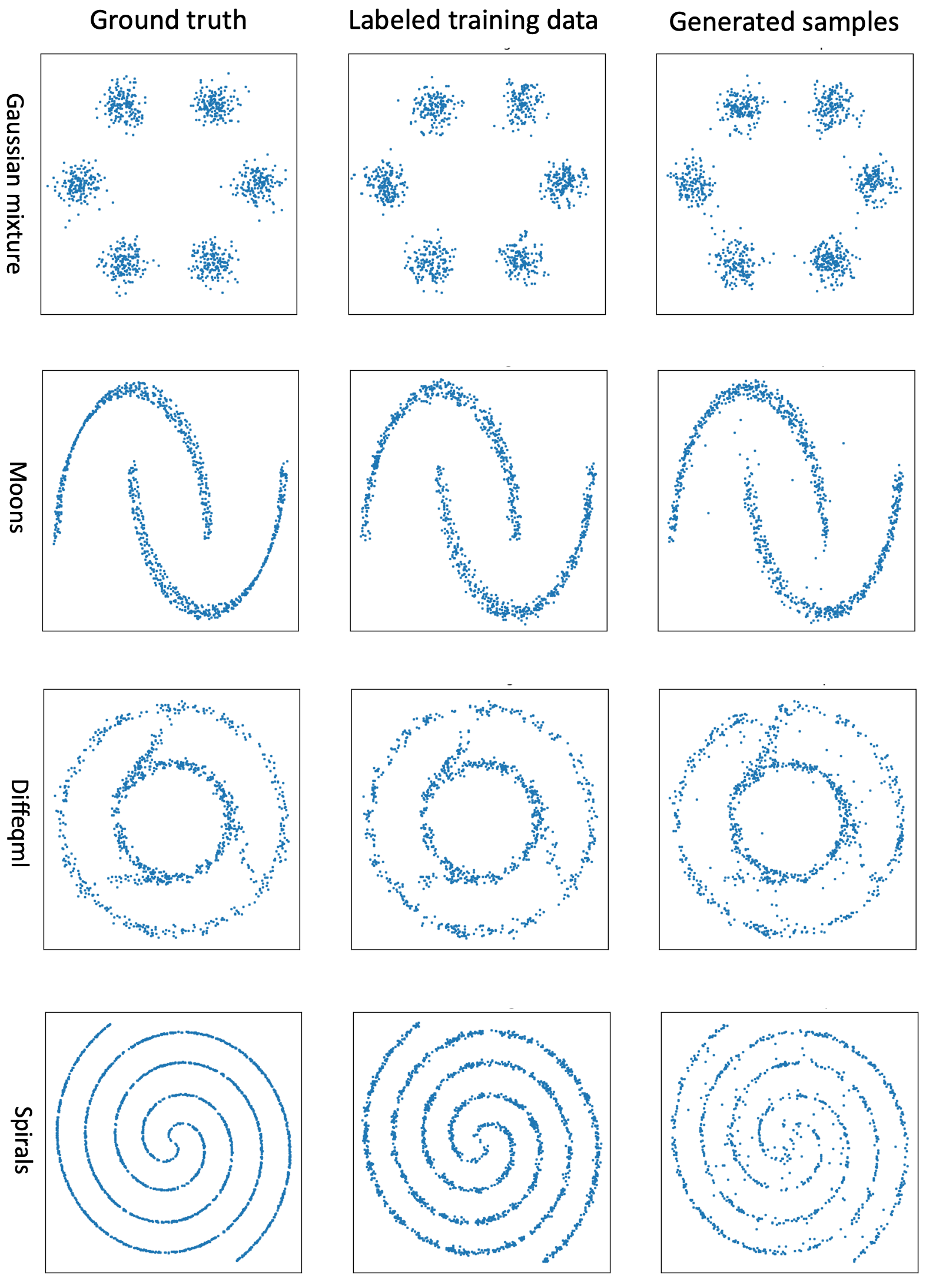}
    \caption{Results on four 2D toy datasets. The left column shows the ground true distribution, i.e., the dataset $\mathcal{X}$; the middle column shows the generated labeled data by solving the ODE in Eq.~\eqref{eq:ode}; and the right column shows the samples generated by the trained generative model $F(y;\theta)$.}
    \label{fig:2d}
\end{figure}

We use four two-dimensional datasets \cite{grathwohl2018ffjord} to demonstrate and visualize the performance of the proposed method. Each dataset has 1000 data, referred to as the ground truth in Figure \ref{fig:2d}.
We use a fully-connected neural network with four hidden layers to define $F(y; \theta)$, each of which has 100 neurons. 100 times steps are used to discretize the reverse-time ODE in Eq.~\eqref{eq:ode} to generate 1000 labeled data. The neural network is then trained with the MSE loss for 5000 epochs using the Adam optimizer with the learning rate chosen as  0.005. 

The results are shown in Figure \ref{fig:2d}. We observe that the labeled samples are not the same as the ground truth,  but they accurately approximate the distribution of the ground truth. In fact, the reverse-time ODE in Eq.~\eqref{eq:ode} can be viewed as a training-free version of the neural-ODE-based normalizing flow \cite{grathwohl2018ffjord}, which can capture multi-modal and discontinuous distributions. The samples generated by $F(y;\theta)$ also provide an accurate approximation to the ground truth, even though the accuracy is lower than the distribution of the labeled training data. There are scattered samples generated by $F(y;\theta)$ because the used neural networks cannot perfectly approximate the discontinuity among different modes of the target distribution.

\subsection{Density Estimation on Real Data}\label{sec:real}
We demonstrate the performance of our method using density estimation on four UCI datasets, including POWER (6-dimensional, available at \small{\url{ http://archive.ics.uci.edu/ml/datasets/Individual+household+electric+power+consumption}}), GAS (8-dimensional, available at \small{\url{ http://archive.ics.uci.edu/ml/datasets/Gas+sensor+array+under+dynamic+gas+mixtures}}), HEPMASS (11-dimensional, available at \small{\url{ http://archive.ics.uci.edu/ml/datasets/HEPMASS}}), and MINIBOONE (43-dimensional, available at \small{\url{ http://archive.ics.uci.edu/ml/datasets/MiniBooNE+particle+identification}}). These datasets are taken from the UCI machine learning repository. We follow the approach in \cite{papamakarios2017masked} to pre-proprocess each dataset. We normalize each dimension of the data by subtracting its mean and dividing its standard deviation, such that all the datasets used to test our method have zero mean and unit standard deviation. Discrete-valued dimensions and every attribute with a Pearson correlation coefficient greater than 0.98 were eliminated.

The reverse-time ODE in Eq.~\eqref{eq:ode} is solved using the explicit Euler scheme with 500 time steps. All training data is split into mini-batches of size $N = 5,000$.  We use a fully-connected neural network with one hidden layer to define $F(y; \theta)$. Specifically, the network for POWER has 256 hidden neurons, the network for GAS has 512 hidden neurons, the network for HEPMASS has 1024 neurons, and the network for MINIBOONE has 1400 hidden neurons, respectively. The neural networks are trained by the Adam optimizer with 20,000 epochs. 

\begin{figure}[h!]
    \centering
    \includegraphics[width=0.85\textwidth]{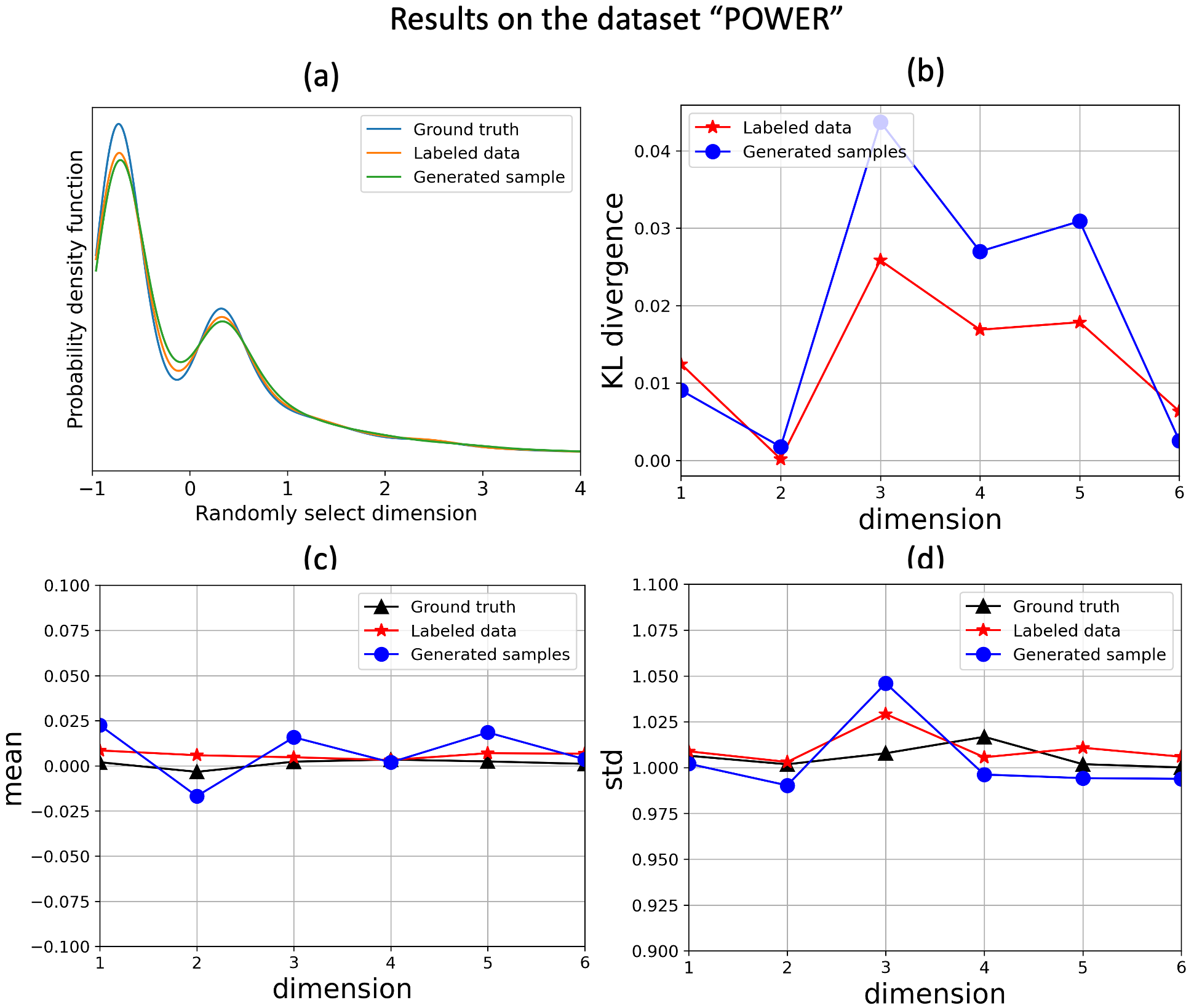}
     \caption{Results on the POWER dataset. (a) The 1D marginal PDF of a randomly selected dimension (b) The K-L divergences of all the 1D marginal distributions; (c) The mean values of all the 1D marginal distributions; (d) The standard deviations of all the 1D marginal distributions.}
    \label{fig3}
\end{figure}

\begin{figure}[h!]
    \centering
    \includegraphics[width=0.85\textwidth]{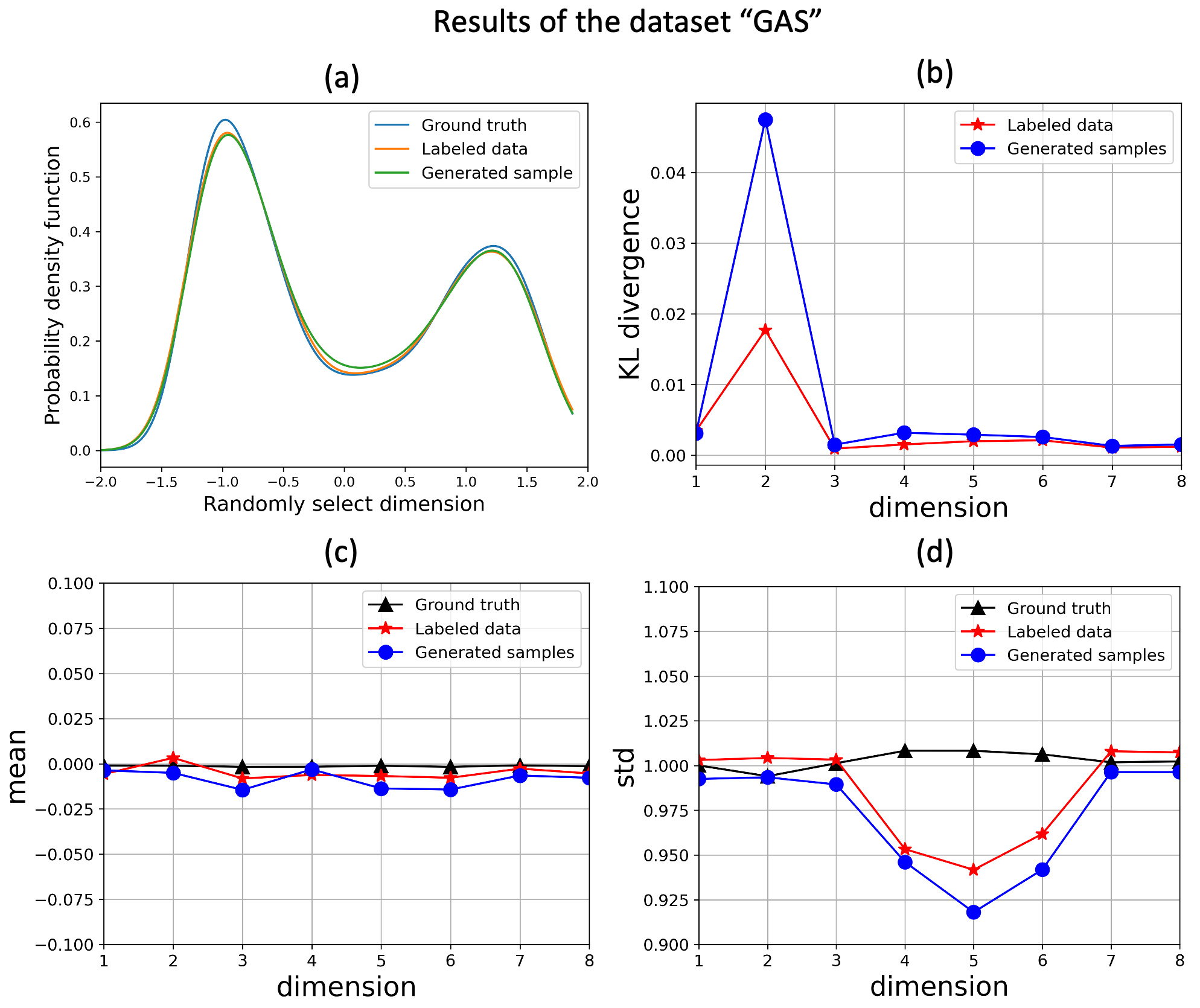}
    \caption{Results on the GAS dataset. (a) The 1D marginal PDF of a randomly selected dimension (b) The K-L divergences of all the 1D marginal distributions; (c) The mean values of all the 1D marginal distributions; (d) The standard deviations of all the 1D marginal distributions.}
    \label{fig4}
\end{figure}

\begin{figure}[h!]
    \centering
    \includegraphics[width=0.85\textwidth]{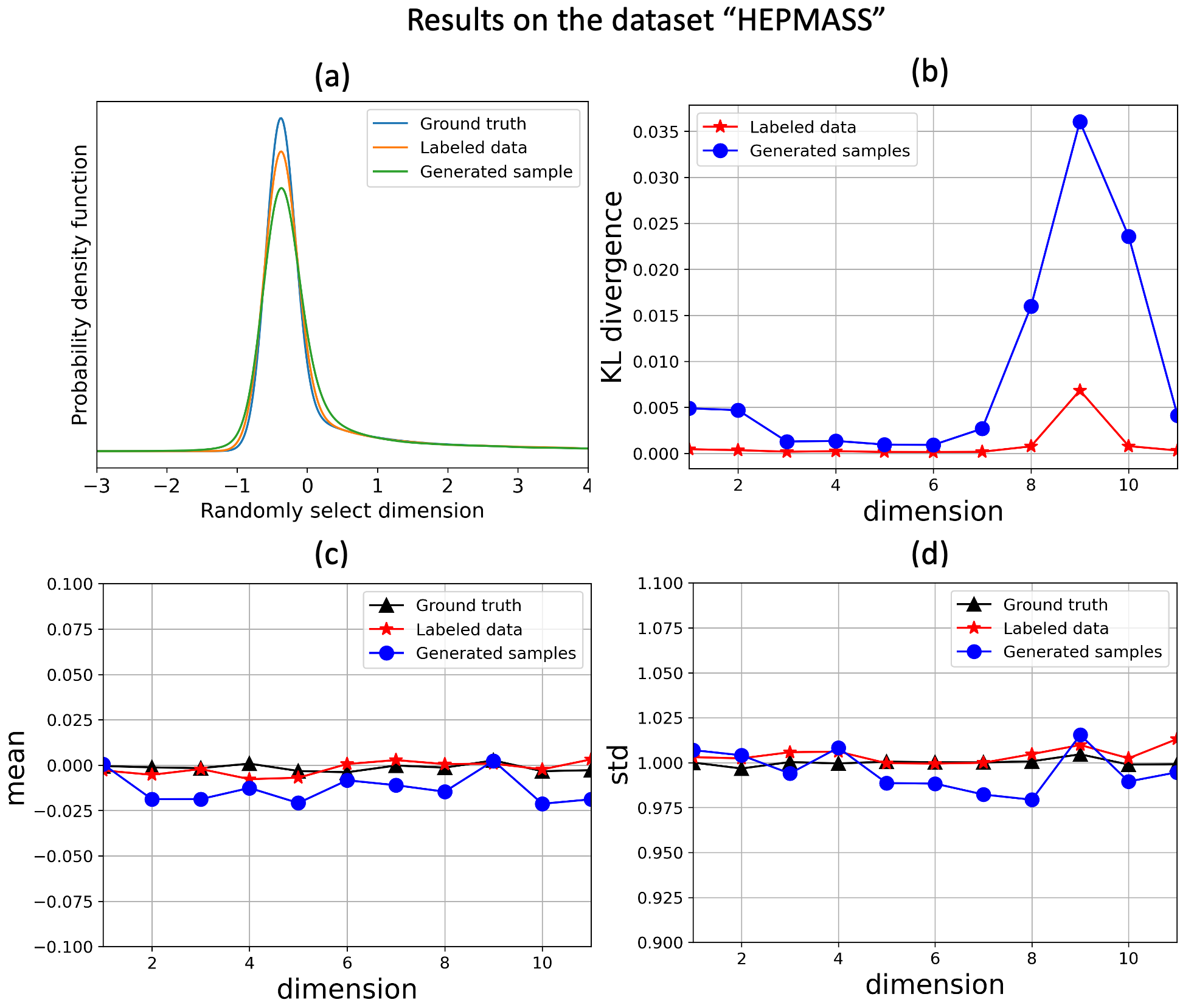}
     \caption{Results on the HEPMASS dataset. (a) The 1D marginal PDF of a randomly selected dimension (b) The K-L divergences of all the 1D marginal distributions; (c) The mean values of all the 1D marginal distributions; (d) The standard deviations of all the 1D marginal distributions.}
    \label{fig5}
\end{figure}

\begin{figure}[h!]
    \centering
    \includegraphics[width=0.85\textwidth]{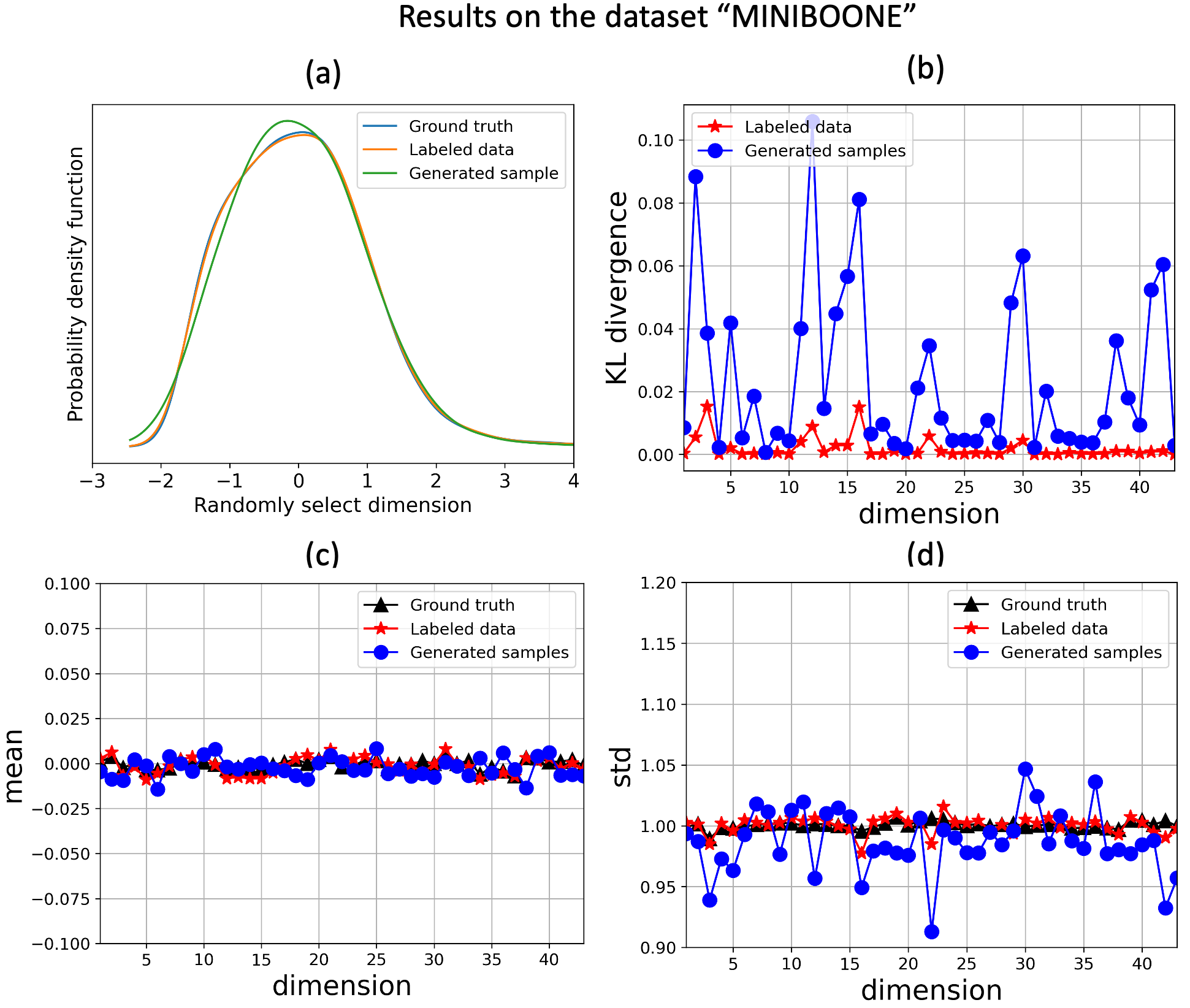}
     \caption{Results on the MINIBOONE dataset. (a) The 1D marginal PDF of a randomly selected dimension (b) The K-L divergences of all the 1D marginal distributions; (c) The mean values of all the 1D marginal distributions; (d) The standard deviations of all the 1D marginal distributions.}
    \label{fig6}
\end{figure}

Figure \ref{fig3} to Figure \ref{fig6} show the comparison among the ground truth data, the labeled data (from the ODE in Eq.~\eqref{eq:ode}) and the generated samples from $F(y;\theta)$ using the following metrics:
\vspace{-0.15cm}
\begin{itemize}\itemsep-0.05cm
   \item The 1D marginal PDF of a randomly selected dimension; 
  \item  The K-L divergences of all the 1D marginal distributions;
  \item The mean values of all the 1D marginal distributions;
  \item The standard deviations of all the 1D marginal distributions.
\end{itemize}
As expected, the labeled data and the generated samples can accurately approximate the distribution of the ground truth. Table \ref{table:NN} shows the computing costs of generating the labeled data, training the neural network $F(y;\theta)$ and using $F(y;\theta)$ to generate samples. The computational time is obtained by running our code on a workstation with Nvidia RTX A5000 GPU. 
\begin{table}[h!]
\centering
\begin{tabular}{c |c c c} 
 \hline
 Dataset  & Data labeling  & Training $F$ & Synthesizing 100K samples \\ 
 \hline
  POWER  & 64.83 seconds & 182.51 seconds & 0.10 seconds\\
  GAS  & 85.78 seconds & 426.61 seconds& 0.23 seconds\\
  HEPMASS  & 109.12 seconds & 1940.87 seconds & 0.51 seconds\\
 MINIBOONE  & 408.36 seconds & 2220.79 seconds& 0.66 seconds\\
 \hline
\end{tabular}
\caption{Wall-clock time of different stages of our method. Data labeling refers to the stage of solving the reverse-time ODE in Eq.~\eqref{eq:ode}; Training $F$ refers to the stage of using the labeled data to train the generative model $F(y;\theta)$; Synthesizing 100K samples refers to using the trained model $F(y;\theta)$ to generate 100K new samples of $X$. We observe that our method features a promising efficiency in generating a large number of samples. }
\label{table:NN}
\end{table}

\section{Conclusion}\label{sec:con}
We introduced a supervised learning framework for training generative models for density estimation. Within this framework, we utilize the score-based diffusion model to generate labeled data and employ  simple,  fully-connected,  neural networks to learn the generative model of interest. The key ingredient is the training-free score estimation that enables data labeling without training the score function. It is important to note that the current algorithm has only been successfully tested using a tabular dataset, and its performance in image/signal synthesis remains to be explored. On the other hand, this algorithm can be applied to a variety of Bayesian sampling problems in scientific and engineering applications, including parameter estimation of physical models, state estimation of dynamical systems (e.g., chemical reactions), surrogate models for particle simulation in physics, all of which will be our future work on this topic.

\section*{Acknowledgement}
This material is based upon work supported by the U.S. Department of Energy, Office of Science, Office of Advanced Scientific Computing Research, Applied Mathematics program under the contract ERKJ387 at the Oak Ridge National Laboratory, which is operated by UT-Battelle, LLC, for the U.S. Department of Energy under Contract DE-AC05-00OR22725. The author Feng Bao would also like to acknowledge the support from the U.S. National Science Foundation through project DMS-2142672 and the support from the U.S. Department of Energy, Office of Science, Office of Advanced Scientific Computing Research, Applied Mathematics program under Grant DE-SC0022297, and the author Yanzhao Cao would also like to acknowledge the support from  the U.S. Department of Energy, Office of Science, Office of Advanced Scientific Computing Research, Applied Mathematics program under Grant DE-SC0022253.

\bibliographystyle{siamplain}
\bibliography{nfref,Reference}

\end{document}